\documentclass[10pt, a4paper]{article}

\usepackage[]{lrec-coling2024} 

\usepackage{algorithm}
\usepackage{algorithmic}
\usepackage{amsfonts,amssymb, amsmath} 
\usepackage{multicol}
\usepackage{multirow}
\usepackage{booktabs}

\DeclareMathOperator*{\argmin}{arg\,min}
\usepackage{color}
\usepackage[misc,geometry]{ifsym}

\title{Unlocking Instructive In-Context Learning with Tabular Prompting for Relational Triple Extraction}

\name{\begin{tabular}{c}Guozheng Li\textsuperscript{$\clubsuit$}, Wenjun Ke\textsuperscript{$\clubsuit \spadesuit$}, Peng Wang\textsuperscript{$\clubsuit \spadesuit$(\Letter)}, Zijie Xu\textsuperscript{$\clubsuit$} \\
Ke Ji\textsuperscript{$\clubsuit$}, Jiajun Liu\textsuperscript{$\clubsuit$}, Ziyu Shang\textsuperscript{$\clubsuit$}, Qiqing Luo\textsuperscript{$\clubsuit$}\end{tabular}}

\address{\textsuperscript{$\clubsuit$}School of Computer Science and Engineering, Southeast University, China \\
\textsuperscript{$\spadesuit$}Key Laboratory of New Generation Artificial Intelligence Technology and Its \\
Interdisciplinary Applications (Southeast University), Ministry of Education, China \\
\{liguozheng, kewenjun, pwang, xuzijie, keji, jiajliu, ziyus1999, qqluo\}@seu.edu.cn\\}

\abstract{
The in-context learning (ICL) for relational triple extraction (RTE) has achieved promising performance, but still encounters two key challenges: (1) how to design effective prompts and (2) how to select proper demonstrations.
Existing methods, however, fail to address these challenges appropriately. 
On the one hand, they usually recast RTE task to text-to-text prompting formats, which is unnatural and results in a mismatch between the output format at the pre-training time and the inference time for large language models (LLMs). 
On the other hand, they only utilize surface natural language features and lack consideration of triple semantics in sample selection. These issues are blocking improved performance in ICL for RTE, thus we aim to tackle prompt designing and sample selection challenges simultaneously.
To this end, we devise a tabular prompting for RTE (\textsc{TableIE}) which frames RTE task into a table generation task to incorporate explicit structured information into ICL, facilitating conversion of outputs to RTE structures. Then we propose instructive in-context learning (I$^2$CL) which only selects and annotates a few samples considering internal triple semantics in massive unlabeled samples. 
Specifically, we first adopt off-the-shelf LLMs to perform schema-agnostic pre-extraction of triples in unlabeled samples using \textsc{TableIE}. 
Then we propose a novel triple-level similarity metric considering triple semantics between these samples and train a sample retrieval model based on calculated similarities in pre-extracted unlabeled data. We also devise three different sample annotation strategies for various scenarios. 
Finally, the annotated samples are considered as few-shot demonstrations in ICL for RTE. 
Experimental results on two RTE benchmarks show that I$^2$CL with \textsc{TableIE} achieves state-of-the-art performance compared to other methods under various few-shot RTE settings.
 \\ \newline \Keywords{Large language models, relation extraction, in-context learning} }

\begin{document}

\maketitleabstract

\section{Introduction}
\begin{figure}[t]
	\centering
	\includegraphics[width=\linewidth]{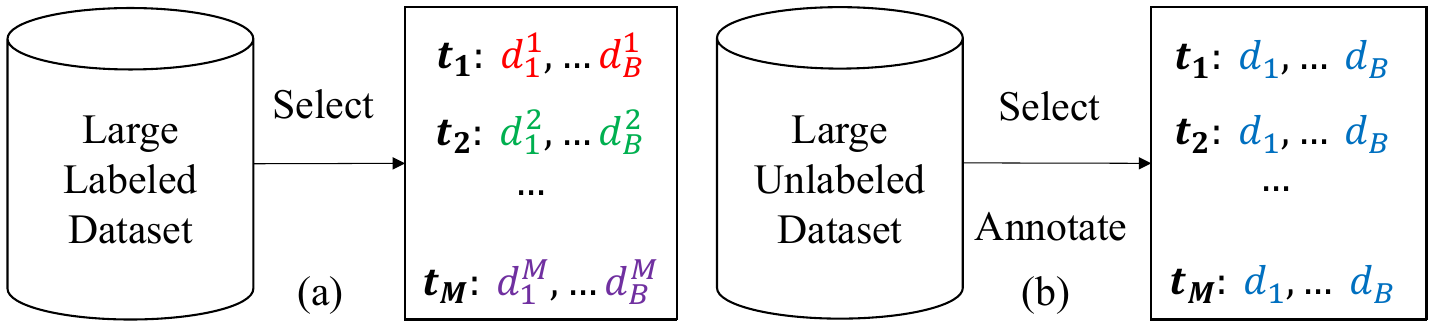}
	\caption{Illustration of the instance-wise retrieving ICL (a) and our I$^2$CL framework (b). For different test samples (e.g., $t_1$ and $t_2$), the former paradigm retrieves $B$ different samples from the large labeled dataset as corresponding demonstrations $d$, while I$^2$CL only selects and annotates a few samples with annotation budget $B$ for all $M$ test samples.}
	\label{icl}
\end{figure}

\begin{figure*}[t]
	\centering
	\includegraphics[width=\linewidth]{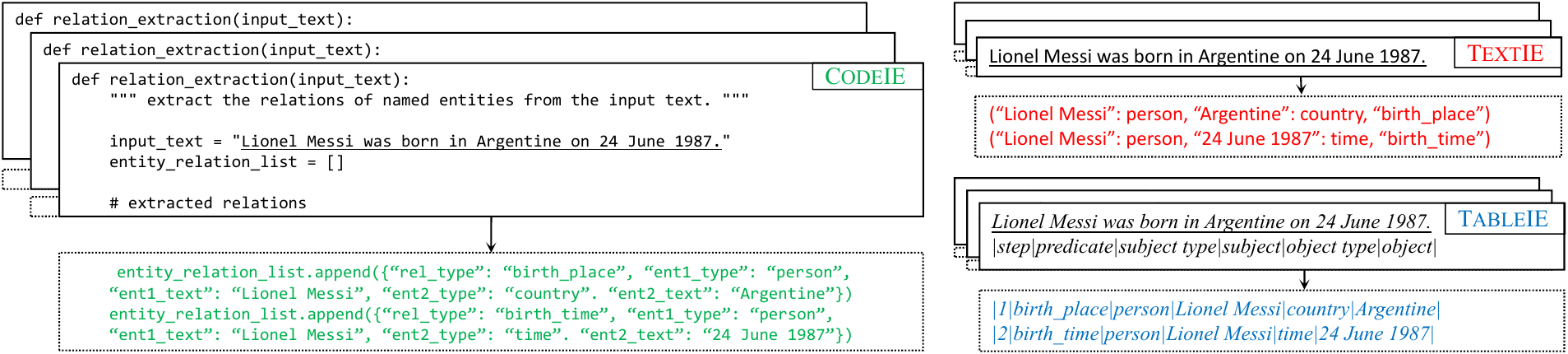}
	\caption{Formats of three prompting. The test sample is marked with \underline{underline}. The outputs of LLMs are highlighted in colors.}
	\label{table}
\end{figure*}

Relational triple extraction (RTE) aims to identify structured information from raw text, involving diverse output structures like named entity recognition (NER)~\cite{zhang2021multi,liu2024crup,wu2024consist} and relation extraction (RE)~\cite{li2022fastre,li2023online,wang2023fmlre,wang2023pascore}. Previous studies~\cite{paolini2021structured, lu2022unified} propose unified frameworks to tackle the RTE task by converting the structured relational triples into unstructured strings and then utilizing text generation models~\cite{lewis-etal-2020-bart, 2020t5}. Recent studies~\cite{wei2022chain, wang2022self,shang2024ontofact} on large-scale pre-trained language models (LLMs), such as GPT-3~\cite{brown2020language}, demonstrate that LLMs perform well in various natural language processing (NLP) tasks without any fine-tuning but only with a few annotated samples as prompts, which is called \textit{in-context learning} (ICL) (shown in Figure~\ref{icl} (a)). 
However, current ICL for RTE still encounters two challenges. On the one hand, recasting RTE task to solely text-to-text prompting format (see Figure~\ref{table} \textsc{TextIE}~\cite{ma2023large}) like other NLP tasks is unnatural and resulting in a mismatch between the output format at the pre-training time and the inference time for LLMs~\cite{li2023codeie}. Therefore, extracting structured data containing multiple dependent elements using text-to-text prompting format makes RTE particularly challenging in ICL.
Thus one important question is \textit{how to design proper prompting formats suitable in ICL for RTE?} 
On the other hand, prior work~\cite{zhao2021calibrate, liu2021makes} has found that the performance of ICL is sensitive to the selected samples. Moreover, due to limited context length of LLMs, only a few annotated samples can be presented in prompts. Therefore, another essential research question is \textit{how to select a few high quality annotated samples as few-shot demonstrations in ICL for RTE?}

Most studies~\cite{lu2022unified, jimenez-gutierrez-etal-2022-thinking, ma2023large, wan2023gpt,liu2024} overlook the first challenge and directly transform the RTE task into text-to-text generation formats like \textsc{TextIE}, producing fragile predicted outputs that require complex decoding strategies for post-processing them into valid structures. \textsc{CodeIE}~\cite{li2023codeie} (see Figure~\ref{table}) argues the abundant structured code information encoded in the pre-trained LLMs can benefit RTE task and achieves superior results compared to \textsc{TextIE}. Despite delivering the significant improvement using code LLMs like Codex~\cite{chen2021evaluating}, its advantage on natural language LLMs like GPT-3 is slightly inferior. Besides, \textsc{CodeIE} inevitably generates more tokens compared to \textsc{TextIE} as shown in Figure~\ref{table}, resulting in higher costs and lower efficiency.
To this end, we devise a tabular prompting for RTE named \textsc{TableIE} that generates organized and concise outputs with lower costs and higher efficiency, incorporating explicit structured table information into ICL. Specifically, a table header ``\textit{$|$step$|$predicate$|$subject type$|$subject$|$object type$|$object}'' is provided as part of the prompt and the LLMs can automatically generate a table, where ``$|$'' is the recognizable delimiter of tables in OpenAI~\footnote{https://openai.com} models. Specially, \textsc{TableIE} is suitable for both zero and few-shot prompting while \textsc{TextIE} and \textsc{CodeIE} can only be applied to few-shot setting as they cannot guarantee the structural integrity under zero-shot prompting in ICL for RTE.

For the second challenge, we argue that two issues need to be considered. 
First, despite retrieving the annotated samples as demonstrations in large labeled dataset (see Figure~\ref{icl} (a)), LLMs ICL still significantly underperforms fine-tuned moderate-size models~\cite{ma2023large}.
Therefore, considering ICL in zero or low-resource rather than high-resource scenarios is more suitable and promising~\cite{su2022selective}. In this work, we formulate an instructive in-context learning (I$^2$CL) framework for RTE (shown in Figure~\ref{icl} (b)): select and annotate a few high-quality samples from the unlabeled data based on the accessible test data, so as to obtain better few-shot prompting results.
Second, raw-input-based sample selection method is one widely applied solution in ICL which involves embedding raw inputs of samples using an off-the-shelf embedding model and then selecting the most similar samples~\cite{rubin2021learning}. Nevertheless, this method is prone to being biased by surface natural language features (syntax, lexicon, semantic, etc.) that may not hold distinct effectiveness for intended tasks~\cite{an2023skill}.
In RTE, raw-input-based selection just finds out samples with similar whole sentence semantics, while the better in-context samples should contain the similar or exactly same entity types and relation types~\cite{wan2023gpt}.

To measure the similarities between samples, features of both entities and relations are required, besides the similarity metrics. 
First, the unlabeled samples contribute limited unsupervised features. Thus we incorporate the strong capability of LLMs on zero-shot prompting~\cite{kojima2022large} to perform schema-agnostic pre-extraction of entities and relations in unlabeled samples using \textsc{TableIE}. The extracted triples is not consistent with annotation schema, but still objectively represent the semantics of the triples contained in samples. 
Second, we propose a novel triple-level similarity metric considering the importance of entity and relation types, describing the similarities between pre-extracted unlabeled samples via Pompeiu–Hausdorff distance~\cite{schutze2012using} because a sample may contain multiple triples. After the above two steps, we obtain the similarities between all unlabeled samples pairwise and then fine-tune a Sentence-BERT~\cite{reimers2019sentence} as sample retriever on the whole similarities calculated unlabeled data, in which the model pays attention to the internal triple semantic differences between samples. During testing, we adopt this model to calculate the similarities between test and unlabeled data.
Then we propose three different sample selection strategies including top-$k$-based, balance-based and coverage-based strategy, where different strategies are suitable for different test data distributions. Finally, we annotate the selected unlabeled samples as demonstrations. In summary, the contributions of our work are three-fold:
\begin{itemize}
    \item We propose a tabular prompting \textsc{TableIE}, incorporating explicit structured table information into ICL that achieves superior performance with lower costs and higher efficiency compared to \textsc{TextIE} and \textsc{CodeIE}. Specially, \textsc{TableIE} can be utilized for both zero and few-shot prompting scenarios in LLMs.
    \item We propose an instructive in-context learning (I$^2$CL) framework, which selects and annotates a few high-quality samples for better prompting results. We propose a novel triple-level similarity metric for instructive sample retrieval and devise three different sample selection strategies suitable for different data distributions.
    \item Experimental results demonstrate that \textsc{TableIE} performs better than \textsc{TextIE} and \textsc{CodeIE}, indicating the advantage of representing structured targets with table. With the same annotation budget, I$^2$CL consistently improves the naive baselines by a notable margin.
\end{itemize}

\section{Related Work}
\paragraph{Zero and Few-shot Prompting} The pre-trained LLMs such as GPT-3~\cite{brown2020language} have demonstrated impressive zero and few-shot learning capabilities across various NLP tasks~\cite{zhao2021calibrate, liu2021makes, wei2022chain, wang2022self, kojima2022large, an2023skill}. Recent studies~\cite{jimenez-gutierrez-etal-2022-thinking, ma2023large, wan2023gpt} on ICL mainly focuses on exploring NER and RE tasks separately, however, there has been relatively little research into potential of LLMs for RTE task~\cite{wei2023zero, li2023codeie}. ChatIE~\cite{wei2023zero} transforms the zero-shot RTE task into a multi-turn question answering problem with a two-stage framework, even surpasses some full shot models on several datasets. \textsc{CodeIE}~\cite{li2023codeie} highlights the beneficial of abundant structured code information encoded in LLMs like Codex~\cite{chen2021evaluating} in information extraction (IE) tasks, delivering superior performance compared to common text-to-text few-shot prompting. To the best of our knowledge, we are the first to devise tabular prompting for RTE.

\paragraph{Sample Selection and Annotation} Prior studies~\cite{zhao2021calibrate, liu2021makes, an2023skill,liu2023iterde} have found that the performance of ICL is sensitive to the selected samples as few-shot demonstrations. In addition, retrieving annotated samples on the large labeled dataset still significantly underperforms fine-tuned models in NER and RE tasks~\cite{jimenez-gutierrez-etal-2022-thinking, ma2023large, wan2023gpt}. Different from the above ICL paradigm, selective annotation~\cite{su2022selective} chooses a pool of samples to annotate from unlabeled data in advance, followed by sample retrieval that retrieves task samples from the annotated pool at test time. However, selective annotation computes the similarities between unlabeled samples using Sentence-BERT~\cite{reimers2019sentence}, where how to select and annotate samples on specific NLP tasks such as RTE is not studied.
Moreover, selective annotation maintains a moderate-size sample pool for subsequent sample retrieval during test phase, while I$^2$CL selects a specific number of samples that are most worth annotating based on the test data. Therefore, I$^2$CL is more suitable for RTE and comes at a lower annotation cost compared to selective annotation.

\paragraph{Active Learning} Our method for ICL shares the same goal of reducing the annotation cost compared to active learning. For example, iterative parameter updates~\cite{wang2016cost,kasai2019low} based active learning methods are computationally expensive for LLMs used in ICL compared to our method. Recently, the effectiveness of active learning has been questioned when LLMs are fine-tuned for various tasks~\cite{karamcheti2021mind}. Our method reduces the annotation cost of ICL, departing from the recent observations on fine-tuning with active learning. The limitations of supervised state-of-the-art models for relation extraction in data-scarce and domain-specific scenarios are discussed in~\citet{mallart2022confronting}, which is similar with our motivation.

\begin{figure*}[t]
	\centering
	\includegraphics[width=\textwidth]{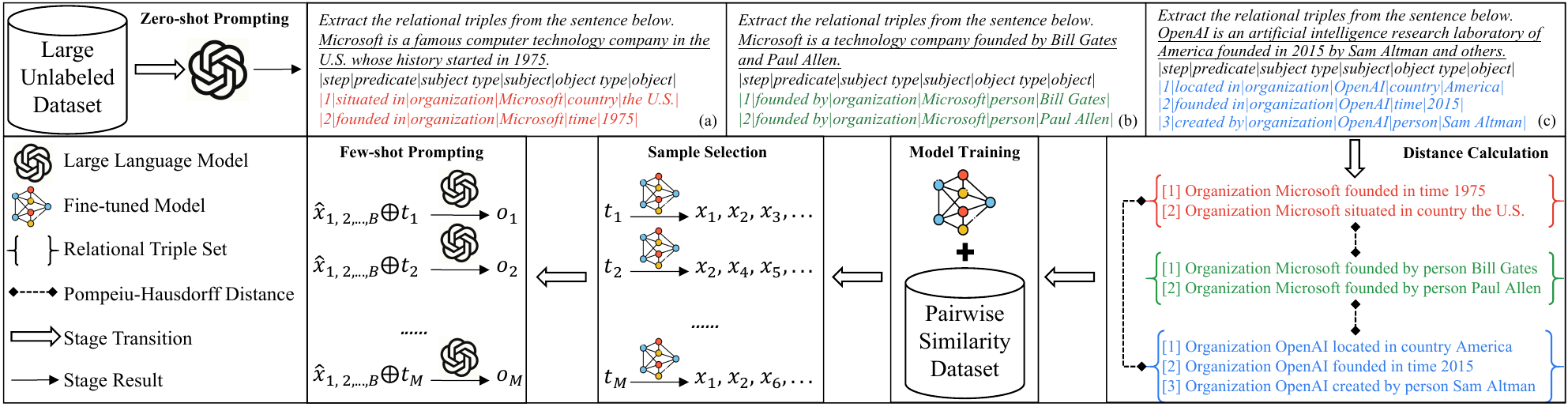}
	\caption{Illustration of the I$^2$CL framework. We aim to measure the similarities of the triple sets contained in two samples and select the most representative samples to annotate based on the whole test data. For instance, sample (b) is very similar to (a) on the surface natural language features while (c) is more similar to (a) in the triple-level semantic features. Moreover, annotate one sample (c) is better than two samples (a) and (b) because (c) contains all the similar triple patterns in (a) and (b).}
	\label{I$^2$CL}
\end{figure*}

\section{Methodology}
\subsection{Problem Statements}
In this work, we assume no large-scale labeled data is available, and we require to select a few samples from existing unlabeled data and annotate them for effective inference. Formally, the objective of I$^2$CL is to select a small subset $\mathcal{S} \subset \mathcal{X}$ from the set of unlabeled samples $\mathcal{X} = \{x_i\}_{i=1}^N$ for annotating contained entities and relations based on the test samples $\mathcal{T} = \{t_i\}_{i=1}^M$, satisfy the annotation budget $B = |\mathcal{S}|$. $N$ and $M$ represent the total number of samples in unlabeled data and test data, respectively. After the selection and annotation, we treat these samples as the few-shot demonstrations in ICL for RTE. We should note that this kind of setting (obtaining the test set in advance to select instances) is practical in real-world scenarios.

\subsection{Framework Overview}
The framework of I$^2$CL is illustrated in Figure~\ref{I$^2$CL}, which consists of five stages: (1) zero-shot prompting, (2) distance calculation, (3) model training, (4) sample selection and (5) few-shot prompting. First, we adopt an off-the-shelf LLM to perform schema-agnostic pre-extraction of entities and relations in the unlabeled samples using \textsc{TableIE}. Second, we calculate the similarities between these unlabeled samples based on the pre-extracted triples via average Pompeiu-Hausdorff distance. Third, we train an efficient sample retrieval model in ICL for RTE. During testing, we calculate the similarities between test samples and unlabeled samples. Then we select and annotate $B$ samples by three different strategies and treat them as the few-shot demonstrations. Below we present each stage in detail.

\subsubsection{Zero-shot Prompting} To obtain the impartial triple-level semantic features inside large unlabeled data, we utilize the capability of LLMs on zero-shot learning to extract all the relational triples including predicates, subject types, subjects, object types and objects. Specifically, we have the following \textsc{TableIE} in zero-shot prompting format:
\begin{table}[ht]
\centering\setlength{\tabcolsep}{1mm}
\fontsize{9}{13.8}\selectfont
\begin{tabular}{l}
\textit{Extract the relational triples from the sentence below.} \\
\textit{$<$Sentence$>$} \\
\textit{$|$step$|$predicate$|$subject type$|$subject$|$object type$|$object$|$} \\
\end{tabular}
\end{table}

\noindent Typically, any valid zero-shot prompting format is compatible with this stage. Unfortunately, existing prompting such as \textsc{TextIE} and \textsc{CodeIE} can only be applied to few-shot prompting, where they are unable to provide precise instructed signals similar to table header in \textsc{TableIE} for RTE. 

\subsubsection{Distance Calculation} After pre-extracting, we need to measure the similarities between the unlabeled samples on pre-extracted triple sets level. Specifically, we first transform each triple containing predicate $p$, subject type $s_t$, subject $s$, object type $o_t$ and object $o$ as the natural language form $z$:
\begin{equation}
    z = s_t \oplus s \oplus p \oplus o_t \oplus o
\end{equation}
where $\oplus$ indicates simple concatenation of strings. The intuition behind our method is that the types of entities and relations are important manifestations of triplet-level semantics, rather than just considering the span of entities. Formally, we define the distance between the triples $z_i$ and $z_j$ as:
\begin{equation}
    d(z_i, z_j) = ||{\rm {Encoder}}(z_i) - {\rm {Encoder}}(z_j)||_2
\end{equation}
where ${\rm {Encoder(\cdot)}}$ denotes the encoder-only pre-trained language model such as Sentence-BERT~\cite{reimers2019sentence} and $||\cdot||_2$ denotes the Euclidean distance. 
Suppose we have two samples with triple sets $\mathcal{Z}_i$ and $\mathcal{Z}_j$, we aim to obtain the Pompeiu-Hausdorff distance between them to measure the similarities by considering all the relational triples simultaneously. As the standard Pompeiu-Hausdorff distance is highly sensitive to outliers, we use the average Pompeiu-Hausdorff distance~\cite{schutze2012using}:
\begin{equation}
    \begin{aligned}
    D(\mathcal{Z}_i, \mathcal{Z}_j) =& \frac{1}{|\mathcal{Z}_i|} \sum_{z_i \in \mathcal{Z}_i} \min_{z_j \in \mathcal{Z}_j} d(z_i, z_j) \, + \\
    & \frac{1}{|\mathcal{Z}_j|} \sum_{z_j \in \mathcal{Z}_j} \min_{z_i \in \mathcal{Z}_i} d(z_i, z_j)
    \end{aligned}
\end{equation}
where the triple sets distance $D(\mathcal{Z}_i, \mathcal{Z}_j)$ serves as the distance of two unlabeled samples in RTE task. 

\subsubsection{Model Training} After obtaining the similarities between all unlabeled samples pairwise, we aim to train an efficient sample retriever for test samples in ICL for RTE. Apparently, we are able to extract the entities and relations from each test sample via LLMs as before, and measure the similarities between unlabeled samples and test samples for selective annotations. However, it is very inconvenient and costly to extract triples from each test sample in advance when new test data arrives, especially with a large number of test samples. To this end, we train an encoder-based model $f_\theta(\cdot)$ which considers solely surface natural language features as inputs but pays more attention to learn the internal relational triple patterns between different samples. Specifically, we fine-tune a Sentence-BERT~\cite{reimers2019sentence} on the similarities calculated unlabeled data which aims to approximately obtain the distance between samples $x_i$ and $x_j$ via:
\begin{equation}
    \mathcal{L} = \sum_{(x_i, x_j) \in \mathcal{X}}(\mathcal{D}(\mathcal{Z}_i, \mathcal{Z}_j) - ||f_\theta(x_i) - f_\theta(x_j)||_2)^2
\end{equation}
where $f_\theta(x)$ denotes the sentence representation of unlabeled sample $x$. We minimize the differences between the values of Pompeiu-Hausdorff distances and sentence representation distances. During testing, we adopt the fine-tuned model to calculate the distances between each test sample and unlabeled sample so as to obtain the pairwise distance set $\mathcal{P} = \{p_{(i,j)}\}_{(i, j)=(1, 1)}^{N \times M}$ between unlabeled samples $\mathcal{X}$ and test samples $\mathcal{T}$:
\begin{equation}
    \mathcal{P} = \{p_{(i,j)} \,| \, p_{(i, j)} = d(x_i, t_j), x_i \in \mathcal{X}, t_j \in \mathcal{T}\}
\end{equation}
where $d(x_i, t_j) = ||f_\theta(x_i) - f_\theta(t_j)||_2$.

\subsubsection{Sample Selection} In sample selection, we select and annotate samples from unlabeled data with annotation budget $B$. We propose three different sample selection strategies based on the pairwise distance set $\mathcal{P}$. 
\begin{itemize}
    \item \textbf{Top-$k$ based strategy} We select the most similar $u$ unlabeled samples for each test sample $t$, resulting in a total of $M \times u$ unlabeled samples. Then we count the frequency of each unlabeled sample occurrence and sort it in descending order of frequency. Finally, we directly select and annotate the first $k$ unlabeled samples for ICL. In this case, we have $k = B$.
    \item \textbf{Balance-based strategy} The top-$k$ method is simple but may cause imbalanced labeling, especially when $B$ is relatively small, which may result in a lack of samples of some specific entity or relation types. If the test data contains $R$ relation types, we select and annotate $B$ samples with $R$-way $\lfloor B/R \rfloor$-shot style in the sorted unlabeled samples. Note this strategy possibly increases the annotation cost, especially under the adverse effects of imbalanced unlabeled data distribution in relation types.
    \item \textbf{Coverage-based strategy} The above two strategies cannot guarantee that the selected unlabeled samples are similar to all the test samples. We thus propose a coverage-based sample selection strategy that ensure the participation of all test samples in selective annotation process, which is described in Algorithm~\ref{alg:1}. We first select the nearest $\lceil M/B \rceil$ test samples for each unlabeled sample $x$ based on pairwise distance set $\mathcal{P}$ (line 1-8). Then the unlabeled sample $x_b$ with the minimum sum of pairwise distances is selected to annotate (line 9-10). Subsequently, we discard the unlabeled sample $x_b$ and its nearest $\lceil M/B \rceil$ test samples in pairwise distance set $\mathcal{P}$ (line 11-12). We iterate through the above process until all the test samples are discarded.
\end{itemize}

\begin{algorithm}[htb!]
\small
    \renewcommand{\algorithmicrequire}{\textbf{Input:}}
	\renewcommand{\algorithmicensure}{\textbf{Output:}}
	\caption{Coverage-based strategy for sample selection}
	\label{alg:1}
	\begin{algorithmic}[1]
		\REQUIRE The unlabeled samples $\mathcal{X} = \{x_i\}_{i=1}^N$, test samples $\mathcal{T} = \{t_i\}_{i=1}^M$ and pairwise distance set $\mathcal{P} = \{p_{(i, j)}\}_{(i, j)=(1, 1)}^{N \times M}$.
		\ENSURE The selective annotation subset $\mathcal{S} \subset \mathcal{X}$ with annotation budget $B = |\mathcal{S}|$.
		\STATE $\mathcal{S} = \varnothing$
		\FOR{$s \leftarrow 1$ to $B$}{
        \IF{$\forall t_j \in \mathcal{T}$, $p_{(*, j)} \notin \mathcal{P}$}
        \STATE \textbf{return} $\mathcal{S}$
        \ELSE
        \FOR{$x_i \in \mathcal{X}$, $p_{(i, *)} \in \mathcal{P}$}{
        \STATE $\mathcal{P}_i = \{p_{(i, *)} \, | \, p_{(i, *)} \in {\rm{min}}(\mathcal{P}, \lceil M/B \rceil)\}$
        }
        \ENDFOR
        \STATE $x_b = \argmin_{x_i \in \mathcal{X}} \sum_{*}^{\lceil M/B \rceil} p_{(i, *)}$ where $p_{(i, *)} \in \mathcal{P}_i$
        \STATE $\mathcal{S} \leftarrow \mathcal{S} \cup x_b$
        \STATE Discard $p_{(i, *)}$ in $\mathcal{P}$ where $x_i = x_b$
        \STATE Discard $p_{(*, j)}$ in $\mathcal{P}$ where $p_{(b, j)} \in \mathcal{P}_b$
        \ENDIF
		}
		\ENDFOR
	\end{algorithmic} 
\end{algorithm}

\subsubsection{Few-shot Prompting} After selecting and annotating $\mathcal{S} = \{(x_i, y_i)\}_{i=1}^B$, we adopt standard few-shot prompting for ICL in RTE. We convert the samples in $\mathcal{S}$ to corresponding table-style pairs $\{(\hat{x}_i, \hat{y}_i)\}_{i=1}^B$. Then we concatenate them as a string to compose the in-context demonstrations $\hat{x}_{1, ...B} = I \oplus \hat{x}_1 \oplus \hat{y}_1 ...\hat{x}_B \oplus \hat{y}_B$, where $I$ denotes the prompt instructions as \textit{Extract the relational triples from the sentences below.} Note that we arrange these demonstrations in ascending order of similarities to the test samples $\mathcal{T}$. Given a specific test sample $t_i$, after feeding the constructed input into the LLMs, we are expected to get an output that is formatted as the same as $y_1, ..., y_B$ which retains the integral table structural: 
\begin{equation}
    o_i = {\rm{LLMs}}\,(\hat{x}_{1,...,B} \oplus t_i)
\end{equation}
where $\hat{x}_{1,...,B} \oplus t_i$ is the prompt for each test sample $t_i \in \mathcal{T}$ and $o_i$ is the extracted relational triples. 

\section{Experiments}

\subsection{Experimental Design}
\subsubsection{Datasets} We evaluate our proposed \textsc{TableIE} and I$^2$CL in RTE task with benchmarks CoNLL04~\cite{roth2004linear} and NYT~\cite{riedel2010modeling} where Table~\ref{ds} shows the dataset statistics. We follow~\citet{lu2022unified} and~\citet{li2023codeie} to preprocess all these datasets. Then we remove all annotated labels from the training set and treat them as unlabeled data. We select and annotate the fake unlabeled training set based on its test set in each dataset.

\begin{table}[h]
\centering\setlength{\tabcolsep}{0.8mm}
\caption{Statistics of the datasets. \# Ents and \# Rels denote the number of entity types and relation types. \# Train, \# Valid and \# Test denote the sample number in each split.}
\begin{tabular}{lccccc}
\toprule
{Dataset} & {\# Ents} & {\# Rels} & {\# Train} & {\# Valid} & {\# Test}\\
\midrule
{CoNLL04} & {4} & {5} & {922} & {231} & {288} \\
{NYT} & {3} & {24} & {56,196} & {5,000} & {5,000} \\
\bottomrule
\end{tabular}
\label{ds}
\end{table}

\subsubsection{Settings} For prompting formats of \textsc{TextIE}~\cite{ma2023large}, \textsc{CodeIE}~\cite{li2023codeie} and our \textsc{TableIE}, we use the same backbone of LLMs including the variant of GPT-3~\cite{brown2020language} ``\texttt{text-davinci-003}'', ChatGPT~\footnote{https://openai.com/blog/chatgpt} ``\texttt{gpt-3.5-turbo}'' and GPT-4~\cite{openai2023gpt4} ``\texttt{gpt-4}''. Note that \textsc{CodeIE} performs well on Codex~\cite{chen2021evaluating}, but unfortunately Codex is deprecated by OpenAI. Typically, we get model predictions by querying OpenAI API in few-shot prompting manner. The hyper-parameters of LLMs with three promptings are keep consistent. 


In I$^2$CL, we train the sample retriever $f_\theta(\cdot)$ for 5 epochs with batch size 16 and learning rate 2e-5 on the pre-extracted unlabeled data using the AdamW~\cite{loshchilov2017decoupled} optimizer. And 10\% of samples in unlabeled data are regarded as the validation data to select the best model $f_\theta(\cdot)$.
For the annotation budget $B$, we experiment with 5, 15 and 25 on CoNLL04, respectively. Due to the context length limit, we are unable to experiment above LLMs with a large budget. Thus we adopt the ``\texttt{gpt-3.5-turbo-16k}'' model to support larger budget 24, 48 and 72 on NYT, respectively. With the same budget, we experiment with our proposed three sample selection strategies: \texttt{top-k} ($u$ = 5), \texttt{balance} and \texttt{coverage} with proposed \textsc{TableIE}. For \textsc{TextIE} and \textsc{CodeIE}, we randomly sampling $k$ training samples for each relation type to construct a $k$-shot demonstration set. We also test the \textsc{TableIE} with randomly selected samples and replace the sample retriever $f_\theta(\cdot)$ with original Sentence-BERT~\footnote{\texttt{all-mpnet-base-v2}}~\cite{reimers2019sentence} to verify the effectiveness of I$^2$CL framework. 

Following previous work~\cite{lu2022unified, li2023codeie}, we use the relation strict F1 as the evaluation metrics for the results of RTE. Specifically, a relational triple prediction is correct only if the relation type is correct and the corresponding offsets and types of its entities are correct. Due to the high variance of few-shot prompting with the random selection method, we conduct three runs with different random seeds for each experiment and report the mean values.

\subsection{Main Results}
As shown in Table~\ref{conll04}, \textsc{TableIE} outperforms other prompting formats and equipping I$^2$CL with \textsc{TableIE} in various LLMs (GPT-3, ChatGPT and GPT-4) consistently achieve superior performance over typical ICL under few-shot settings on CoNLL04, demonstrating the effectiveness of our proposed \textsc{TableIE} prompting and I$^2$CL framework. 
On the one hand, \textsc{TableIE} outperforms \textsc{TextIE} under different experimental settings and achieves competitive or superior performance compared to \textsc{CodeIE}, which highlights the importance and beneficial of incorporating explicit structure information into RTE task. 
On the other hand, I$^2$CL with \texttt{balance} and \texttt{coverage} delivers significantly improvement compared to \textsc{TableIE} with random selection, which indicates that appropriate sample selection strategies provide instructive suggestions for annotating representative samples in ICL for RTE. Concretely, \texttt{balance} and \texttt{coverage} achieve similar performance while the latter is suitable for larger annotation budgets. However, \texttt{top-k} fails to achieve promising results on CoNLL04 especially when the annotation budget is relatively small. We provide a concrete example to analyze this phenomenon. Intuitively, \texttt{top-k} is highly affected by the distribution of test data relation types, where 47 relational triples in CoNLL04 test data belongs to relation \textit{Kill} and 105 belongs to \textit{OrgBased\_In}. Empirically, we discover that there are no relational triples belonging to \textit{Kill} with annotated budget 5. With the increase of annotated budget, the adverse effects brought by \texttt{top-k} have been significantly alleviated, but it is still inferior to random selection on CoNLL04. Moreover, when using the same kind of prompting and comparing the used LLMs, \texttt{gpt-4} demonstrates stronger extraction capability than the other two LLMs \texttt{text-davinci-003} and \texttt{gpt-3.5-turbo}, but the performance gap between the three is not notable, especially when experiment with less annotated data and random selection strategy. Generally, the sample selection strategies improve final ICL performance by 4.59\% on \texttt{text-davinci-003} and 6.74\% on \texttt{gpt-4} compared to \texttt{random}. Without sample selection, only 0.44\%, 0.91\% and 0.61\% improvement gained compared to \texttt{text-davinci-003} for \texttt{gpt-4} with $B$ = 5, 15 and 25, respectively. Thus we argue that I$^2$CL provide instructive sample annotation and treat these representative samples as few-shot demonstrations stimulates LLMs stronger capability of understanding relational triples.

\begin{table}[t]
\centering\setlength{\tabcolsep}{0.5mm}
\caption{Experimental results on CoNLL04 benchmark. Best results with different budgets and models are in \textbf{bold}.}
\begin{tabular}{l|l|lll}
\toprule
\multirow{2}{*}{\textbf{Model}} & \multirow{2}{*}{\textbf{Method}} & \multicolumn{3}{c}{\textbf{CoNLL04}} \\
{} & {} & {$B$=5} & {$B$=15} & {$B$=25} \\
\midrule
\multirow{6}{*}{GPT-3} & {\textsc{TextIE}} & {19.85} & {32.83} & {40.35} \\

{} & {\textsc{CodeIE}} & {36.23} & {41.29} & {49.98} \\

{} & {\textsc{TableIE}} & {\textbf{37.36}} & {\textbf{42.90}} & {\textbf{50.77}} \\
\cmidrule{2-5}
{} & + {I$^2$CL}$_{\texttt{top-k}}$ & {25.97} \textcolor{red}{$\downarrow$} & {39.49} \textcolor{red}{$\downarrow$} & {47.71} \textcolor{red}{$\downarrow$} \\
{} & + {I$^2$CL}$_{\texttt{balance}}$ & {\textbf{39.30}} \textcolor{blue}{$\uparrow$} & {\textbf{46.77}} \textcolor{blue}{$\uparrow$} & {54.97} \textcolor{blue}{$\uparrow$} \\
{} & + {I$^2$CL}$_{\texttt{coverage}}$ & {36.21} \textcolor{blue}{$\uparrow$} & {45.41} \textcolor{blue}{$\uparrow$} & {\textbf{55.36}} \textcolor{blue}{$\uparrow$} \\
\midrule
\multirow{6}{*}{ChatGPT} & {\textsc{TextIE}} & {23.55} & {37.53} & {42.95} \\

{} & {\textsc{CodeIE}} & {36.83} & {42.87} & {49.78} \\

{} & {\textsc{TableIE}} & {\textbf{37.29}} & {\textbf{43.15}} & {\textbf{50.31}} \\
\cmidrule{2-5}
{} & + {I$^2$CL}$_{\texttt{top-k}}$ & {27.70} \textcolor{red}{$\downarrow$} & {43.56} \textcolor{blue}{$\uparrow$} & {49.72} \textcolor{red}{$\downarrow$} \\
{} & + {I$^2$CL}$_{\texttt{balance}}$ & {\textbf{40.18}} \textcolor{blue}{$\uparrow$} & {47.24} \textcolor{blue}{$\uparrow$} & {55.33} \textcolor{blue}{$\uparrow$} \\
{} & + {I$^2$CL}$_{\texttt{coverage}}$ & {36.35} \textcolor{blue}{$\uparrow$} & {\textbf{47.38}} \textcolor{blue}{$\uparrow$} & {\textbf{56.45}} \textcolor{blue}{$\uparrow$} \\
\midrule
\multirow{6}{*}{GPT-4} & {\textsc{TextIE}} & {24.92} & {38.35} & {46.53} \\

{} & {\textsc{CodeIE}} & {36.77} & {43.25} & {50.19} \\

{} & {\textsc{TableIE}} & {\textbf{37.80}} & {\textbf{43.81}} & {\textbf{51.38}} \\
\cmidrule{2-5}
{} & + {I$^2$CL}$_{\texttt{top-k}}$ & {28.41} \textcolor{red}{$\downarrow$} & {44.11} \textcolor{blue}{$\uparrow$} & {50.44} \textcolor{red}{$\downarrow$} \\
{} & + {I$^2$CL}$_{\texttt{balance}}$ & {\textbf{40.51}} \textcolor{blue}{$\uparrow$} & {\textbf{48.93}} \textcolor{blue}{$\uparrow$} & {57.54} \textcolor{blue}{$\uparrow$} \\
{} & + {I$^2$CL}$_{\texttt{coverage}}$ & {37.63} \textcolor{blue}{$\uparrow$} & {48.70} \textcolor{blue}{$\uparrow$} & {\textbf{58.12}} \textcolor{blue}{$\uparrow$} \\
\bottomrule
\end{tabular}
\label{conll04}
\end{table}

\begin{table}[t]
\centering\setlength{\tabcolsep}{0.5mm}
\caption{Experimental results on NYT benchmark. Best results with different budgets and models are in \textbf{bold}. The results are all based on the \texttt{gpt-3.5-turbo-16k}.}
\begin{tabular}{l|l|lll}
\toprule
\multirow{2}{*}{\textbf{Model}} & \multirow{2}{*}{\textbf{Method}} & \multicolumn{3}{c}{\textbf{NYT}} \\
{} & {} & {$B$=24} & {$B$=48} & {$B$=72} \\
\midrule
\multirow{6}{*}{ChatGPT} & {\textsc{TextIE}} & {18.85} & {18.88} & {19.44} \\
{} & {\textsc{CodeIE}} & {28.23} & {28.75} & {29.78} \\
{} & {\textsc{TableIE}} & \textbf{29.31} & \textbf{29.84} & \textbf{30.45} \\
\cmidrule{2-5}
{} & + {I$^2$CL}$_{\texttt{top-k}}$ & \textbf{{32.28}} \textcolor{blue}{$\uparrow$} & \textbf{{34.92}} \textcolor{blue}{$\uparrow$} & \textbf{{35.66}} \textcolor{blue}{$\uparrow$} \\
{} & + {I$^2$CL}$_{\texttt{balance}}$ & {29.23} \textcolor{red}{$\downarrow$} & {30.24} \textcolor{blue}{$\uparrow$}& {30.42} \textcolor{red}{$\downarrow$}\\
{} & + {I$^2$CL}$_{\texttt{coverage}}$ & {31.44} \textcolor{blue}{$\uparrow$} & {32.21} \textcolor{blue}{$\uparrow$} & {32.89} \textcolor{blue}{$\uparrow$}\\
\bottomrule
\end{tabular}
\label{nyt}
\end{table}

We further compare these approaches under different annotation budget on NYT. As shown in Table~\ref{nyt}, we can see that the discovered phenomenons on CoNLL04 still hold except for the differences in performance among the three sample selection strategies. We notice that \texttt{top-k} can achieve much better performance compared to its counterparts especially with larger annotation budget, while \texttt{balance} only deliver similar results compared to random selection. Here we provide the empirical analysis. Specifically, NYT is an extremely imbalanced dataset where the imbalance of samples belonging to different relations makes it difficult to select samples with appropriate proportions of different relations in ICL. Note that \textit{/location/location/contains} is the most frequently appearing relation in NYT test data, where total 3,827 of 8,110 relational triples belonging to this relation. In contrast, some relations such as \textit{/people/person/ethnicity} and \textit{/people/ethnicity/geographic\_distribution} correspond to very few samples (i.e., only one sample) in whole test data. When selecting demonstrations in ICL for such extremely imbalanced test data, it is unreasonable to sample $k$ samples for each relation to construct a $k$-shot demonstration set. Typically, we cannot know in advance the proportion of each relation in the test set, and using \texttt{balance} may not necessarily achieve good results. Therefore, we should choose the best strategy based on the distribution of test data, where \texttt{coverage} is stabler and better than other two strategies and consistently achieves satisfactory results with insensitivity to test data distribution.

\begin{figure}[t]
\includegraphics[width=0.49\linewidth]{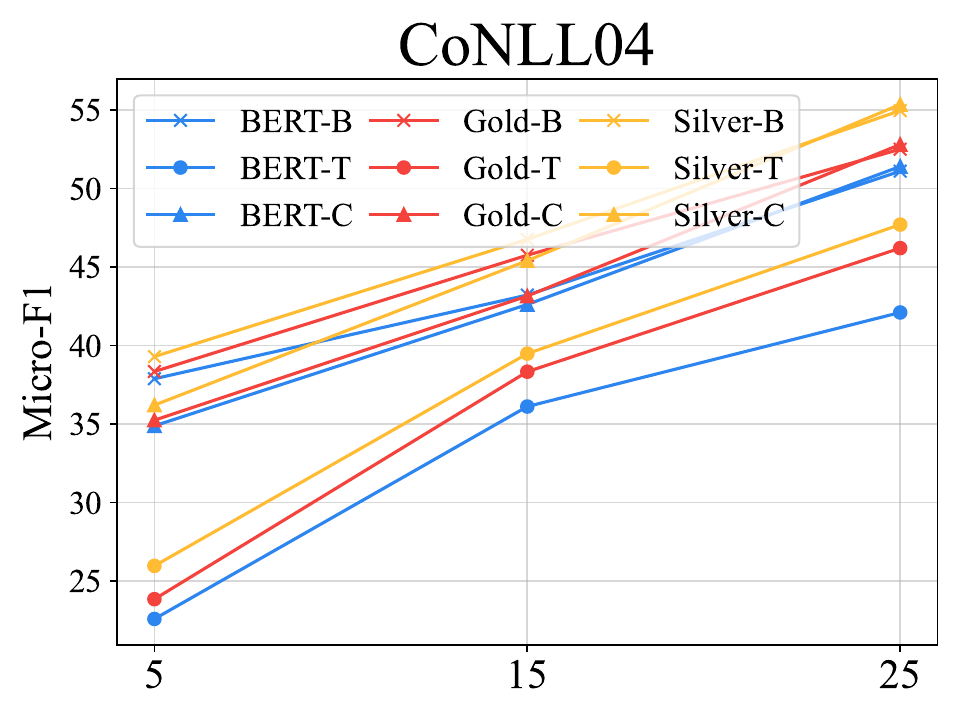} \label{c} \includegraphics[width=0.49\linewidth]{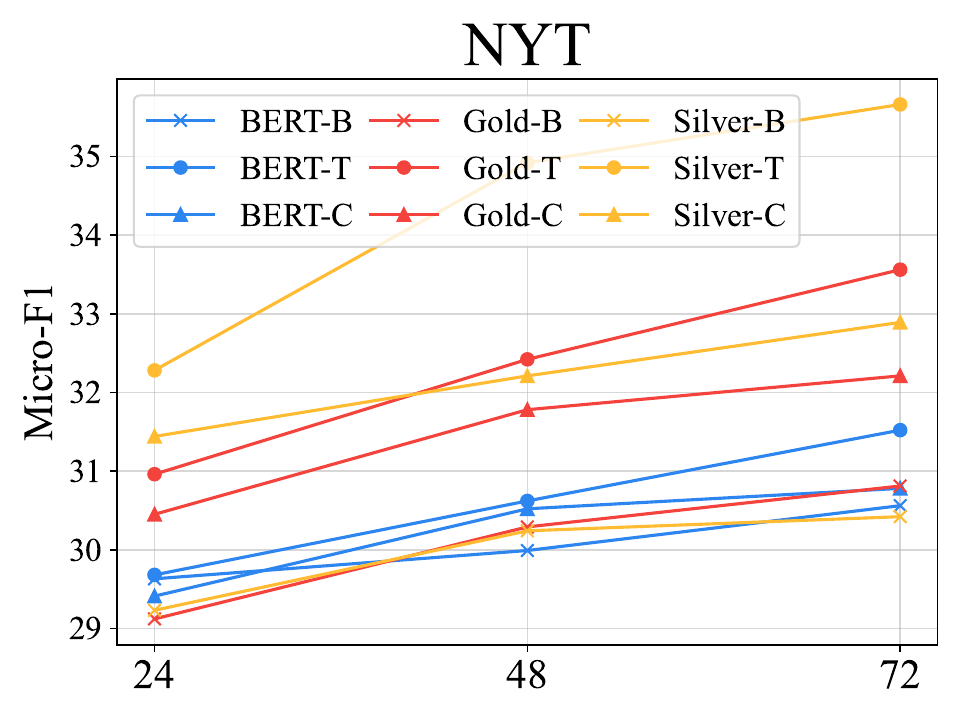} \label{n}
\caption{Performance of different retrieval models on two benchmarks. We use \texttt{text-davinci-003} on CoNLL04 and \texttt{gpt-3.5-turbo-16k} on NYT. BERT denotes the original Sentence-BERT without fine-tuning. Gold denotes the fine-tuned model on annotated training data, while Silver denotes the fine-tuned model on training data with pre-extraction results. And B, T, C represent \texttt{balance}, \texttt{top-k}, and \texttt{coverage} strategies, respectively.}
\label{retrieval}
\end{figure}

\subsection{Different Retrieval Results}
To verify the effectiveness of retrieval model in I$^2$CL, we evaluate the impact of different retrieval models on final ICL performance. Besides the proposed retrieval model (Silver), we also adopt the original Sentence-BERT model (BERT) and train a new retrieval model (Gold) based on the original annotated training data. As shown in Figure~\ref{retrieval}, considering relational triple features and adopting Pompeiu-Hausdorff distance as similarity metric bring better retrieving results. Consistent with the conclusion of the main experiment, \texttt{balance} and \texttt{coverage} perform better on CoNLL04 while \texttt{top-k} is more suitable for NYT. Intuitively, retrieval model based on ground truth would achieve the best results. However, Gold is only slightly better than BERT and worse than Silver on two benchmarks. This counter-intuitive phenomenon is likely due to two reasons. One is that there are many samples with incorrect annotation or incomplete annotation in the training set of datasets, especially the NYT which is completely constructed by distant supervision~\cite{mintz2009distant}. These inaccurate noise signals can degrade the final performance of retrieval model. Another is that we force the model to learn to retrieve the samples that similar to test samples in triple-level semantics from the perspective of LLMs rather than annotators. Compared to previous approaches, our metric serves as a more accurate proxy for evaluating the utility of a training sample during testing. 
In contrast, BERT solely retrieves samples with similar sentence semantics but ignore the subtle semantics of entities and relations inside samples, where the improvement of using BERT as retrieval model is modest compared to random selection without any retriever.

\begin{table}[t]
\centering\setlength{\tabcolsep}{0.1mm}
\caption{Model generalization results on CoNLL04 and NYT. Best results are in \textbf{bold}. Results that have improved compared to the original ones are marked with \underline{underline}.}
\begin{tabular}{l|ccc|ccc}
\toprule
\multirow{2}{*}{\textbf{Method}} & \multicolumn{3}{c}{\textbf{CoNLL04}} & \multicolumn{3}{c}{\textbf{NYT}}\\
{} & {$B$=5} & {$B$=15} & {$B$=25} & {$B$=24} & {$B$=48} & {$B$=72}\\
\midrule
{\textsc{TableIE}} & {37.36} & {42.90} & {50.77} & {29.31} & {29.84} & {30.45} \\
+ {I$^2$CL}$_{\texttt{top-k}}$ & \underline{{28.97}} & \underline{{40.71}} & \underline{{49.31}} & {31.36} & \textbf{{32.62}} & \textbf{{33.79}} \\
+ {I$^2$CL}$_{\texttt{balance}}$ & \textbf{{37.88}} & \textbf{{44.37}} & {53.12} & {29.10} & {30.14} & {30.20} \\
+ {I$^2$CL}$_{\texttt{coverage}}$ & {35.78} & {43.40} & \textbf{{53.26}} & \textbf{{31.43}} & {31.56} & \underline{{33.21}} \\
\bottomrule
\end{tabular}
\label{ability}
\end{table}

\subsection{Model Generalization Ability}
In this section we explore the retrieval generalization ability. 
Specifically, we use the retrieval model trained on unlabeled data of NYT on CoNLL04, simulating cases where a retrieval model is directly deployed in a new scenario to test model generalization. We are able to experiment in this manner because the pre-extraction process using zero-shot prompting is schema-agnostic. In other words, we expect the retrieval model to learn the objectively semantic information and can be adapted to new datasets. We conduct experiments with \texttt{text-davinci-003} on CoNLL04 and \texttt{gpt-3.5-turbo} on NYT both with \textsc{TableIE}, showing results in Table~\ref{ability}. With the exchangeable retrieval model, the improvement from random selection to strategic selection is less than 3\%, indicating the data drift issue degrading the model performance to some extent. Under different budget settings, however, I$^2$CL still enjoys obvious advantages regardless of diverse LLMs as backbones. Note that the exchangeable retrieval model also improves the performance of \texttt{top-k} on CoNLL04 and \texttt{coverage} on NYT, which indicates that our method achieves moderate generalization ability across datasets.


\begin{table}[t]
\centering\setlength{\tabcolsep}{1.0mm}
\caption{Costs and efficiency results. \# Total denotes the total number of characters in LLMs outputs, and \# Avg., \# Min. and \# Max. represent the average, minimum and maximum values of character lengths of all samples, respectively.}
\begin{tabular}{l|cccc}
\toprule
{\textbf{Method}} & {\textbf{\# Total}} $\downarrow$ & {\textbf{\# Avg.}} $\downarrow$ & {\textbf{\# Min.}} $\downarrow$ & {\textbf{\# Max.}} $\downarrow$ \\
\midrule
{\textsc{TextIE}} & {28,473} & {98.86} & {38} & {527} \\
{\textsc{CodeIE}} & {71,431} & {248.02} & {131} & {2456} \\
{\textsc{TableIE}} & \textbf{24,976} & \textbf{86.72} & \textbf{30} & \textbf{460} \\
\bottomrule
\end{tabular}
\label{cost}
\end{table}

\subsection{Costs and Efficiency}
Specially, we compare the costs and efficiency of three different prompting formats using \texttt{text-davinci-003} on CoNLL04. The experimental results are illustrated in Table~\ref{cost}. Since we are unable to access the specific tokenizer in LLMs, we estimated the characters generated using different methods. Generally, generating more characters (i.e., more tokens) means spending more response time and consuming more computing resources, suffering from very low efficiency. Specifically, the total number of characters generated using \textsc{CodeIE} is around 2.9 $\times$ as \textsc{TableIE}, which is not surprising because it requires to generate redundant tokens such as the keys in Python dictionaries. Moreover, the minimum and maximum values of character lengths is 4.37 - 5.34 $\times$ as \textsc{TableIE}. Compared with \textsc{TextIE}, our \textsc{TableIE} only generates results with less than 0.88 $\times$ average character length, but achieves significantly better performance in ICL for RTE, demonstrating the superior of \textsc{TableIE}.

\subsection{Discussions on Settings}
Obviously, the proposed method needs to obtain the test set in advance to select instances. We acknowledge that the setting defined in this work is not very common. Generally, we will assume that the test samples are unknown, and then design corresponding algorithms to select ICL samples in advance, as defined in previous work~\cite{su2022selective}. However, in a lot of practical scenarios, selecting and annotating samples in advance is so ideal and typically brings suboptimal results. 

For example, imagine we are developing a system to extract relational triples from news articles. We start with a set of unlabeled news articles. These articles cover various topics, including politics, sports, and entertainment. But we are aware that the model may not perform well on specific and niche topics that are not well-represented in the training data (e.g., the new test samples belong to military topic). As new articles are published, we receive a stream of unlabeled test samples. These test samples may cover emerging events, new personalities, or unique scenarios not present in the initial training data. Instead of pre-annotating a fixed set of demonstrations, we dynamically annotate a few examples from each batch of test samples based on their topics. These annotations serve as in-context demonstrations.

A very important question is: why do we have to choose unchanging samples in advance for ICL when targeting future unknown test samples? We argue that a reasonable solution is selecting then annotating a few good samples as demonstrations for specific test samples in a specific scenario and time. Annotating samples in advance has potential risks: we do not know in advance what entity and relation types test samples will contain, which can lead to pre-annotated samples deviating from test samples in terms of type distribution. In our setting, training a triple-level similarity retriever in a specific domain is valuable, because we can provide personalized demonstrations for every time new test samples in this domain arrive compared to unchanging samples. Specifically, the setting in ~\citet{su2022selective} works well with small distribution deviation between test samples and annotated samples. Our setting expects to determine the distribution of annotated samples based on test samples. In summary, we argue that in a lot of practical scenarios, it is reasonable and sometimes necessary to annotate a few high quality samples with less human labor for potentially massive test samples to be predicted, considering the performance requirements in real-world scenarios.

\subsection{Beyond GPT-Series Models}
Besides the OpenAI GPT-Series LLMs, we also consider other LLMs such as LLaMA~\cite{touvron2023llama}, T5~\cite{2020t5}, OPT~\cite{zhang2022opt} etc. However, we empirically find these open-source LLMs basically cannot perform zero and few-shot prompting in RTE similar with the findings in~\citet{li-etal-2023-revisiting-large} perhaps due to the LLMs are not large enough to perform ICL. For example, we tried to experiment LLaMA-7B on RTE task with several demonstrations via few-shot prompting paradigm. However, we observe that LLaMA cannot understand the instructions and is unable to recover the expected triples based on demonstrations, and LLaMA just starts to completely repeat the input sentences. Current ICL research in information extraction (IE) only focuses on very large models~\cite{li2023codeie,ma2023large,li-etal-2023-revisiting-large} because smaller foundation models are unable to perform ICL in IE just like simpler NLP tasks such as sentiment analysis.

\subsection{Why \textsc{TableIE} Works}
Foremost, TEXIE and CODEIE cannot guarantee the structural integrity under zero-shot prompting because there is no explicit prompt (i.e., table header) to guide them in generating valid triples without few-shot demonstrations. Since prompting is a brittle process wherein small modifications to the prompt can cause large variations in the model predictions, we are unable to theoretically judge the effectiveness of a prompting other than through empirical results. But we can provide some relevant inspiring works to enhance our conclusion. As the output of RTE is structured, incorporating explicit structured information into ICL tends to benefit the final RTE performance~\cite{li2023codeie}. Besides, reasoning and extracting answers step by step in such tabular format is effective in other downstream tasks~\cite{ziqi2023tab}. And LLMs such as GPT-3~\cite{brown2020language} and CodeX~\cite{chen2021evaluating} have the capability of reasoning over tabular structured data, because such models are trained on massive tabular formed data. 

\section{Conclusion}
In this work, we devise a tabular prompting \textsc{TableIE}, framing the RTE task to a table generation task and achieving promising performance in ICL. We propose I$^2$CL, an instructive in-context learning framework for RTE, which is more effective and realistic in zero or low-resource scenarios. I$^2$CL leverages the capability of LLMs on zero-shot prompting and natural language understanding to achieve better few-shot prompting results with less annotation. We also propose a novel triple-level similarity metric for sample retrieval. Besides, three sample selection strategies are proposed to annotate proper samples with a few annotation budget, where we may require to choose the best strategy based on data distribution according to empirical results, encouraging more effective methods in the future research.

\section*{Acknowledgements}
We thank the reviewers for their insightful comments. This work was supported by National Science Foundation of China (Grant Nos.62376057) and the Start-up Research Fund of Southeast University (RF1028623234).All opinions are of the authors and do not reflect the view of sponsors.

\nocite{*}
\section{Bibliographical References}\label{sec:reference}

\bibliographystyle{lrec-coling2024-natbib}
\bibliography{lrec-coling2024-example}


\appendix

\section{Retrieval with Pompeiu-Hausdorff}
We experiment the strategy (we call this strategy “Z+S”) of Zero-shot prompting (extracting triples from test samples) + Sentence-BERT (using Sentence-BERT to select unlabeled samples via Pompeiu-Hausdorff similarities), and this strategy could achieve similar or sometimes slightly better results (but typically less than 1\%) compared to our fine-tuned retriever, as shown in Table~\ref{zs_conll04} and Table~\ref{zs_nyt}. This is reasonable because the aim of fine-tuned retriever is to approximate the selection quality of this strategy but without extra pre-extraction on possibly massive test samples as mentioned in Model Training section. In other words, the fine-tuned retriever is expected to select good demonstrations taking solely surface natural language features as inputs but pays more attention to the internal relational triple patterns during testing.

\begin{table}[ht]
\centering\setlength{\tabcolsep}{0.5mm}
\caption{Experimental results on CoNLL04 benchmark. Best results with different budgets and models are in \textbf{bold}.}
\begin{tabular}{l|l|lll}
\toprule
\multirow{2}{*}{\textbf{Model}} & \multirow{2}{*}{\textbf{Method}} & \multicolumn{3}{c}{\textbf{CoNLL04}} \\
{} & {} & {$B$=5} & {$B$=15} & {$B$=25} \\
\midrule
\multirow{9}{*}{GPT-3} & {\textsc{TextIE}} & {19.85} & {32.83} & {40.35} \\

{} & {\textsc{CodeIE}} & {36.23} & {41.29} & {49.98} \\

{} & {\textsc{TableIE}} & {\textbf{37.36}} & {\textbf{42.90}} & {\textbf{50.77}} \\
\cmidrule{2-5}
{} & + {I$^2$CL}$_{\texttt{top-k}}$ & {25.97} \textcolor{red}{$\downarrow$} & {39.49} \textcolor{red}{$\downarrow$} & {47.71} \textcolor{red}{$\downarrow$} \\
{} & + {Z+S}$_{\texttt{top-k}}$ & {25.30} \textcolor{red}{$\downarrow$} & {40.22} \textcolor{red}{$\downarrow$} & {48.64} \textcolor{red}{$\downarrow$} \\
{} & + {I$^2$CL}$_{\texttt{balance}}$ & {39.30} \textcolor{blue}{$\uparrow$} & {46.77} \textcolor{blue}{$\uparrow$} & {54.97} \textcolor{blue}{$\uparrow$} \\
{} & + {Z+S}$_{\texttt{balance}}$ & {\textbf{39.38}} \textcolor{blue}{$\uparrow$} & {\textbf{46.96}} \textcolor{blue}{$\uparrow$} & {55.60} \textcolor{blue}{$\uparrow$} \\
{} & + {I$^2$CL}$_{\texttt{coverage}}$ & {36.21} \textcolor{blue}{$\uparrow$} & {45.41} \textcolor{blue}{$\uparrow$} & {55.36} \textcolor{blue}{$\uparrow$} \\
{} & + {Z+S}$_{\texttt{coverage}}$ & {36.42} \textcolor{blue}{$\uparrow$} & {45.03} \textcolor{blue}{$\uparrow$} & {\textbf{55.88}} \textcolor{blue}{$\uparrow$} \\
\midrule
\multirow{9}{*}{ChatGPT} & {\textsc{TextIE}} & {23.55} & {37.53} & {42.95} \\

{} & {\textsc{CodeIE}} & {36.83} & {42.87} & {49.78} \\

{} & {\textsc{TableIE}} & {\textbf{37.29}} & {\textbf{43.15}} & {\textbf{50.31}} \\
\cmidrule{2-5}
{} & + {I$^2$CL}$_{\texttt{top-k}}$ & {27.70} \textcolor{red}{$\downarrow$} & {43.56} \textcolor{blue}{$\uparrow$} & {49.72} \textcolor{red}{$\downarrow$} \\
{} & + {Z+S}$_{\texttt{top-k}}$ & {27.24} \textcolor{red}{$\downarrow$} & {43.50} \textcolor{blue}{$\uparrow$} & {50.48} \textcolor{blue}{$\uparrow$} \\
{} & + {I$^2$CL}$_{\texttt{balance}}$ & {40.18} \textcolor{blue}{$\uparrow$} & {47.24} \textcolor{blue}{$\uparrow$} & {55.33} \textcolor{blue}{$\uparrow$} \\
{} & + {Z+S}$_{\texttt{balance}}$ & {\textbf{40.40}} \textcolor{blue}{$\uparrow$} & {47.31} \textcolor{blue}{$\uparrow$} & {56.10} \textcolor{blue}{$\uparrow$} \\
{} & + {I$^2$CL}$_{\texttt{coverage}}$ & {36.35} \textcolor{blue}{$\uparrow$} & {47.38} \textcolor{blue}{$\uparrow$} & {\textbf{56.45}} \textcolor{blue}{$\uparrow$} \\
{} & + {Z+S}$_{\texttt{coverage}}$ & {36.42} \textcolor{blue}{$\uparrow$} & {\textbf{47.67}} \textcolor{blue}{$\uparrow$} & {56.37} \textcolor{blue}{$\uparrow$} \\
\midrule
\multirow{9}{*}{GPT-4} & {\textsc{TextIE}} & {24.92} & {38.35} & {46.53} \\

{} & {\textsc{CodeIE}} & {36.77} & {43.25} & {50.19} \\

{} & {\textsc{TableIE}} & {\textbf{37.80}} & {\textbf{43.81}} & {\textbf{51.38}} \\
\cmidrule{2-5}
{} & + {I$^2$CL}$_{\texttt{top-k}}$ & {28.41} \textcolor{red}{$\downarrow$} & {44.11} \textcolor{blue}{$\uparrow$} & {50.44} \textcolor{red}{$\downarrow$} \\
{} & + {Z+S}$_{\texttt{top-k}}$ & {28.75} \textcolor{red}{$\downarrow$} & {44.39} \textcolor{blue}{$\uparrow$} & {51.04} \textcolor{red}{$\downarrow$} \\
{} & + {I$^2$CL}$_{\texttt{balance}}$ & {40.51} \textcolor{blue}{$\uparrow$} & {48.93} \textcolor{blue}{$\uparrow$} & {57.54} \textcolor{blue}{$\uparrow$} \\
{} & + {Z+S}$_{\texttt{balance}}$ & {\textbf{40.62}} \textcolor{blue}{$\uparrow$} & {49.45} \textcolor{blue}{$\uparrow$} & {58.50} \textcolor{blue}{$\uparrow$} \\
{} & + {I$^2$CL}$_{\texttt{coverage}}$ & {37.63} \textcolor{blue}{$\uparrow$} & {48.70} \textcolor{blue}{$\uparrow$} & {58.12} \textcolor{blue}{$\uparrow$} \\
{} & + {Z+S}$_{\texttt{coverage}}$ & {37.98} \textcolor{blue}{$\uparrow$} & {\textbf{49.64}} \textcolor{blue}{$\uparrow$} & {\textbf{58.53}} \textcolor{blue}{$\uparrow$} \\
\bottomrule
\end{tabular}
\label{zs_conll04}
\end{table}

\begin{table}[t]
\centering\setlength{\tabcolsep}{0.5mm}
\caption{Experimental results on NYT benchmark. Best results with different budgets and models are in \textbf{bold}. The results are all based on the \texttt{gpt-3.5-turbo-16k}.}
\begin{tabular}{l|l|lll}
\toprule
\multirow{2}{*}{\textbf{Model}} & \multirow{2}{*}{\textbf{Method}} & \multicolumn{3}{c}{\textbf{NYT}} \\
{} & {} & {$B$=24} & {$B$=48} & {$B$=72} \\
\midrule
\multirow{9}{*}{ChatGPT} & {\textsc{TextIE}} & {18.85} & {18.88} & {19.44} \\
{} & {\textsc{CodeIE}} & {28.23} & {28.75} & {29.78} \\
{} & {\textsc{TableIE}} & \textbf{29.31} & \textbf{29.84} & \textbf{30.45} \\
\cmidrule{2-5}
{} & + {I$^2$CL}$_{\texttt{top-k}}$ & {32.28} \textcolor{blue}{$\uparrow$} & {34.92} \textcolor{blue}{$\uparrow$} & {35.66} \textcolor{blue}{$\uparrow$} \\
{} & + {Z+S}$_{\texttt{top-k}}$ & \textbf{{33.12}} \textcolor{blue}{$\uparrow$} & \textbf{{35.64}} \textcolor{blue}{$\uparrow$} & \textbf{{36.27}} \textcolor{blue}{$\uparrow$} \\
{} & + {I$^2$CL}$_{\texttt{balance}}$ & {29.23} \textcolor{red}{$\downarrow$} & {30.24} \textcolor{blue}{$\uparrow$}& {30.42} \textcolor{red}{$\downarrow$}\\
{} & + {Z+S}$_{\texttt{balance}}$ & {29.74} \textcolor{blue}{$\uparrow$} & {30.80} \textcolor{blue}{$\uparrow$}& {30.67} \textcolor{blue}{$\uparrow$}\\
{} & + {I$^2$CL}$_{\texttt{coverage}}$ & {31.44} \textcolor{blue}{$\uparrow$} & {32.21} \textcolor{blue}{$\uparrow$} & {32.89} \textcolor{blue}{$\uparrow$}\\
{} & + {Z+S}$_{\texttt{coverage}}$ & {31.48} \textcolor{blue}{$\uparrow$} & {32.94} \textcolor{blue}{$\uparrow$} & {33.30} \textcolor{blue}{$\uparrow$}\\
\bottomrule
\end{tabular}
\label{zs_nyt}
\end{table}

\section{Comparison with Vote-k}
We mentioned the vote-$k$ (selective annotation) in our related work but not explicitly discussed it in our experiments because we tend to think this kind of comparison is neither fair or necessary.

First, vote-$k$ and I$^2$CL are applied in different practical settings. The vote-$k$ selects samples to annotate before test time while I$^2$CL selects samples to annotate when new test samples arrive. In other words, vote-$k$ finds a few global representative samples in advance to annotate but may result in suboptimal results for test samples in the future, while I$^2$CL finds local optimal demonstrations for current test samples.

Second, we empirically find that vote-k obviously underperform I$^2$CL, shown in Table~\ref{new_conll04} and Table~\ref{new_nyt}. But this is not surprising because vote-$k$ is still based on sentence-BERT similarity calculation, ignoring the internal triple semantics in test samples. For example, the improvements brought by vote-$k$ become relatively obvious for GPT-4 compared to \textsc{TableIE} baseline. In text-davinci-3 results, with the same annotation budgets, vote-$k$ improves the vanilla \textsc{TableIE} method by typically less than 0.5\% absolute gain or is even worse than our random selection baseline (e.g., vote-$k$ delivers around 36.42\% (-0.94\%), 42.17\% (-0.73\%) and 51.22\% (+0.45\%) on CoNLL04 with budget $B$=5, 15 and 25, respectively). The samples selected by vote-$k$ lack consideration for entity and relation semantics, resulting in minimal advantage compared to directly sampling $k$ samples for each relation type. Despite vote-$k$ is worse than I$^2$CL in our setting, the purposes and settings of two methods are very different, and this finding does not necessarily demonstrate the superiority of I$^2$CL.

\begin{table}[t]
\centering\setlength{\tabcolsep}{0.5mm}
\caption{Experimental results on CoNLL04 benchmark. Best results with different budgets and models are in \textbf{bold}.}
\begin{tabular}{l|l|lll}
\toprule
\multirow{2}{*}{\textbf{Model}} & \multirow{2}{*}{\textbf{Method}} & \multicolumn{3}{c}{\textbf{CoNLL04}} \\
{} & {} & {$B$=5} & {$B$=15} & {$B$=25} \\
\midrule
\multirow{7}{*}{GPT-3} & {\textsc{TextIE}} & {19.85} & {32.83} & {40.35} \\

{} & {\textsc{CodeIE}} & {36.23} & {41.29} & {49.98} \\

{} & {\textsc{TableIE}} & {\textbf{37.36}} & {\textbf{42.90}} & {\textbf{50.77}} \\
\cmidrule{2-5}
{} & + {vote-$k$} & {36.42} \textcolor{red}{$\downarrow$} & {42.17} \textcolor{red}{$\downarrow$} & {51.22} \textcolor{blue}{$\uparrow$} \\
{} & + {I$^2$CL}$_{\texttt{top-k}}$ & {25.97} \textcolor{red}{$\downarrow$} & {39.49} \textcolor{red}{$\downarrow$} & {47.71} \textcolor{red}{$\downarrow$} \\
{} & + {I$^2$CL}$_{\texttt{balance}}$ & {\textbf{39.30}} \textcolor{blue}{$\uparrow$} & {\textbf{46.77}} \textcolor{blue}{$\uparrow$} & {54.97} \textcolor{blue}{$\uparrow$} \\
{} & + {I$^2$CL}$_{\texttt{coverage}}$ & {36.21} \textcolor{blue}{$\uparrow$} & {45.41} \textcolor{blue}{$\uparrow$} & {\textbf{55.36}} \textcolor{blue}{$\uparrow$} \\
\midrule
\multirow{7}{*}{ChatGPT} & {\textsc{TextIE}} & {23.55} & {37.53} & {42.95} \\

{} & {\textsc{CodeIE}} & {36.83} & {42.87} & {49.78} \\

{} & {\textsc{TableIE}} & {\textbf{37.29}} & {\textbf{43.15}} & {\textbf{50.31}} \\
\cmidrule{2-5}
{} & + {vote-$k$} & {36.87} \textcolor{red}{$\downarrow$} & {42.56} \textcolor{red}{$\downarrow$} & {51.64} \textcolor{blue}{$\uparrow$} \\
{} & + {I$^2$CL}$_{\texttt{top-k}}$ & {27.70} \textcolor{red}{$\downarrow$} & {43.56} \textcolor{blue}{$\uparrow$} & {49.72} \textcolor{red}{$\downarrow$} \\
{} & + {I$^2$CL}$_{\texttt{balance}}$ & {\textbf{40.18}} \textcolor{blue}{$\uparrow$} & {47.24} \textcolor{blue}{$\uparrow$} & {55.33} \textcolor{blue}{$\uparrow$} \\
{} & + {I$^2$CL}$_{\texttt{coverage}}$ & {36.35} \textcolor{blue}{$\uparrow$} & {\textbf{47.38}} \textcolor{blue}{$\uparrow$} & {\textbf{56.45}} \textcolor{blue}{$\uparrow$} \\
\midrule
\multirow{7}{*}{GPT-4} & {\textsc{TextIE}} & {24.92} & {38.35} & {46.53} \\

{} & {\textsc{CodeIE}} & {36.77} & {43.25} & {50.19} \\

{} & {\textsc{TableIE}} & {\textbf{37.80}} & {\textbf{43.81}} & {\textbf{51.38}} \\
\cmidrule{2-5}
{} & + {vote-$k$} & {37.22} \textcolor{red}{$\downarrow$} & {45.16} \textcolor{blue}{$\uparrow$} & {53.25} \textcolor{blue}{$\uparrow$} \\
{} & + {I$^2$CL}$_{\texttt{top-k}}$ & {28.41} \textcolor{red}{$\downarrow$} & {44.11} \textcolor{blue}{$\uparrow$} & {50.44} \textcolor{red}{$\downarrow$} \\
{} & + {I$^2$CL}$_{\texttt{balance}}$ & {\textbf{40.51}} \textcolor{blue}{$\uparrow$} & {\textbf{48.93}} \textcolor{blue}{$\uparrow$} & {57.54} \textcolor{blue}{$\uparrow$} \\
{} & + {I$^2$CL}$_{\texttt{coverage}}$ & {37.63} \textcolor{blue}{$\uparrow$} & {48.70} \textcolor{blue}{$\uparrow$} & {\textbf{58.12}} \textcolor{blue}{$\uparrow$} \\
\bottomrule
\end{tabular}
\label{new_conll04}
\end{table}

\begin{table}[t]
\centering\setlength{\tabcolsep}{0.5mm}
\caption{Experimental results on NYT benchmark. Best results with different budgets and models are in \textbf{bold}. The results are all based on the \texttt{gpt-3.5-turbo-16k}.}
\begin{tabular}{l|l|lll}
\toprule
\multirow{2}{*}{\textbf{Model}} & \multirow{2}{*}{\textbf{Method}} & \multicolumn{3}{c}{\textbf{NYT}} \\
{} & {} & {$B$=24} & {$B$=48} & {$B$=72} \\
\midrule
\multirow{7}{*}{ChatGPT} & {\textsc{TextIE}} & {18.85} & {18.88} & {19.44} \\
{} & {\textsc{CodeIE}} & {28.23} & {28.75} & {29.78} \\
{} & {\textsc{TableIE}} & \textbf{29.31} & \textbf{29.84} & \textbf{30.45} \\
\cmidrule{2-5}
{} & + {vote-$k$} & {30.54} \textcolor{blue}{$\uparrow$} & {31.63} \textcolor{blue}{$\uparrow$} & {31.77} \textcolor{blue}{$\uparrow$} \\
{} & + {I$^2$CL}$_{\texttt{top-k}}$ & \textbf{{32.28}} \textcolor{blue}{$\uparrow$} & \textbf{{34.92}} \textcolor{blue}{$\uparrow$} & \textbf{{35.66}} \textcolor{blue}{$\uparrow$} \\
{} & + {I$^2$CL}$_{\texttt{balance}}$ & {29.23} \textcolor{red}{$\downarrow$} & {30.24} \textcolor{blue}{$\uparrow$}& {30.42} \textcolor{red}{$\downarrow$}\\
{} & + {I$^2$CL}$_{\texttt{coverage}}$ & {31.44} \textcolor{blue}{$\uparrow$} & {32.21} \textcolor{blue}{$\uparrow$} & {32.89} \textcolor{blue}{$\uparrow$}\\
\bottomrule
\end{tabular}
\label{new_nyt}
\end{table}

\section{Clarify of Balanced Strategy}
In balanced strategy, we mention that this strategy possibly increases the annotation cost, especially under the adverse effects of imbalanced unlabeled data distribution in relation types. The potentially higher cost of annotation comes from some of the top sorted unlabeled samples may potentially belong to the same relation type. Suppose we aim to annotate 15 samples for 5 relations (i.e., 3 samples for each relation), but we may find that the top sorted 15 samples only contain 1 or 2 samples for a specific relation. Then we perhaps require the top sorted 20-30 samples to complete this balanced annotation. We call this “higher annotation cost” because even though we only annotate 15 samples in the end, we actually checked over 15 samples. But this potential for more checks will not appear in other two strategies.

\section{Clarify of Term "Annotation"}
The “annotation” actually means that the labels of selected samples are annotated by humans in practical scenarios, not LLMs. In the context of this paper, we assume that the labels in the annotated datasets are unknown (in fact, the samples have already been annotated and serve as benchmarks). We perform our \textsc{TableIE} and I$^2$CL methods on these “fake” unlabeled samples, then the selected unlabeled samples are “annotated” by aligning sentences with golden labels in original annotated datasets. In other words, we imitated the process of human annotation of data.

\end{document}